\documentclass{IOS-Book-Article}

\usepackage{mathptmx}
\usepackage{soul}\setuldepth{article}
\usepackage{comment}
\usepackage{enumitem}
\usepackage{amsthm}
\usepackage{quoting}
\usepackage{cleveref}
\usepackage[htt]{hyphenat}
\newtheorem{definition}{Definition}
\usepackage{lipsum}
\usepackage{graphicx}
\graphicspath{ {./images/} }

\def\hb{\hbox to 11.5 cm{}}

\usepackage{listings}

\usepackage{xcolor}

\begin{document}

\pagestyle{headings}
\def\thepage{}
\begin{frontmatter}              


\title{We'll give you what you want: An Ontological Model of User Preferences\thanks{The research reported in this paper has been supported by the German Research Foundation DFG, as part of Collaborative Research Center (Sonderforschungsbereich) 1320 Project-ID 329551904 ``EASE - Everyday Activity Science and Engineering'', University of Bremen (\url{http://www.ease-crc.org/}). The research was conducted in subprojects ``P01 -- Embodied semantics for the language of action and change: Combining
analysis, reasoning and simulation'', ``PO5-N -- Principles of Metareasoning for Everyday Activities'', ``INF -- Research Data Management within EASE and Distribution of Open Research Data'' and by the FET-Open Project \#951846 ''MUHAI – Meaning and Understanding for Human-centric AI'' by the EU Pathfinder and Horizon 2020 Program, and by the trilateral Project \#442588247 ''ai4hri – artificial intelligence for human-robot interaction'' which is partly funded by the DFG.}}

\markboth{}{July 2023\hb}

\author[A]{\fnms{Mona} \snm{Abdel-Keream}},
\author[A]{\fnms{Daniel} \snm{Be\ss ler}},
\author[B]{\fnms{Ayden} \snm{Janssen}},
\author[A]{\fnms{Sascha} \snm{Jongebloed}},
\author[B]{\fnms{Robin} \snm{Nolte}},
\author[B]{\fnms{Mihai} \snm{Pomarlan}},
and
\author[B]{\fnms{Robert} \snm{Porzel}}

\address[A]{University of Bremen, Institute for Artificial Intelligence, Bremen, Germany}
\address[B]{University of Bremen, Department of Linguistics, Bremen, Germany}

\begin{abstract}
The notion of preferences plays an important role in many disciplines including service robotics which is concerned with scenarios in which robots interact with humans. These interactions can be favored by robots taking human preferences into account. This raises the issue of how preferences should be represented to support such preference-aware decision making. Several formal accounts for a notion of preferences exist. However, these approaches fall short on defining the nature and structure of the options that a robot has in a given situation. In this work, we thus investigate a formal model of preferences where options are non-atomic entities that are defined by the complex situations they bring about.
\end{abstract}

\end{frontmatter}
\markboth{July 2023\hb}{July 2023\hb}

\section{Introduction}

Robots have yet to reach the level of cognitive capabilities required for use-cases in the service sector where they interact with humans, for example as shopkeepers, waiters or general assistants.
Through these interactions a course of events ensue that can be more or less suitable to the wishes and intentions of the human.
The robot may face different options that would each bring about a different situation where some individual outcome might be preferable over others from the perspective of the human.
However, the actual preference of humans over different options in a given situation is highly context-dependent.
A human might, for example, prefer cold over hot drinks on summer days, while preferring hot drinks like tea and coffee when eating a cake.

The concept of preferences is relevant in many disciplines ranging from philosophy, economics and social science to robotics.
A unified understanding of preferences across those disciplines has however yet to emerge.
A significant step towards this goal by means of a logical language is the framework of preference logic~\cite{hansson2001preference}.
It defines several fundamental characteristics such as asymmetry of preference, and allows to formulate statements that relate different options.
However, the structure of options and their relations to the situations they bring about is widely unconsidered.

In this work we, thus, seek to endow our cognitive robotic agents with the conceptual inventory needed to reason about the possible preferences of the involved human agents in the light of the resulting situation for a given action path. 
To this end, we introduce a formal model of preferences that is purposed as a means for reasoning and ultimately selecting a specific course of actions over alternative ones with respect to the given user preference at hand.
The model is designed along a set of competency questions that each corresponds to a robot capability required in the scope of human-robot interaction scenarios.
We use the framework of description logics for the formalization of the model which has the advantage of having a trivial implementation in the \emph{Web Ontology Language (OWL)}.
The resulting ontology implementation is then used for the validation of the model by showcasing that the competency questions can be implemented in form of queries within a standard language tool.

Summarizing, the contributions of this work are the following ones:
\begin{itemize}
\item a formal ontological model of user preferences where options are characterized through situations they bring about; and 
\item the alignment of this model with an existing robotics domain ontology.
\end{itemize}


\section{Related Work}
Taking user preferences into account in an agent's decision-making system is fundamental for ensuring an enhanced user experience \cite{canal} \cite{hellou}. By considering these preferences, the system can suggest items that align with the user's desires while filtering out items they might dislike. Recommendation systems, also known as recommender systems, leverage user preferences to offer highly personalized recommendations across various domains, including items, products, content, or service.

There exist several types of recommender systems, each employing different techniques to achieve personalized suggestions. Collaborative Filtering \cite{abdo8} relies on analyzing the preferences and behaviors of similar users to recommend items. In contrast, Content-Based Filtering \cite{lim}  takes into account attributes of items and recommends items with similar attributes to a user based on past interactions. Both filtering methods suffer from the so-called cold-start problem \cite{Schein} which arises when the recommendation system lacks sufficient data about a new user or item to generate meaningful suggestions. Knowledge-Based Filtering \cite{alir} goes further by incorporating explicit knowledge about user preferences and item characteristics into their recommendation process. By adopting a knowledge-based approach, the need for extensive data sets is eliminated due to the utilization of domain knowledge as the foundation \cite{bouraga}.  Hybrid recommender systems \cite{mart2019} are commonly employed by combining these filtering techniques, resulting in more robust and reliable recommendations. All previously cited work considers preferences among objects, which is sufficient for e-commerce applications but, we argue, not for robotics where the ontological status of preferences is more complex -- as we argue in section~\ref{sec:ontology}, it is necessary to consider hypothetical situations, not just objects.

According to \cite{Adomavicius2011} three types of knowledge need to be modeled, including : user knowledge, item knowledge and the contextual knowledge relating the item to the user’s needs. This information can be modeled using ontologies. A  robust ontological model is the SOMA model. SOMA provides a structured framework for effectively capturing and representing the contextual knowledge that influence human actions and interactions across diverse everyday situations.

In \cite{CAO2011162}  , a  model is proposed to bridge the product design gap between customers' unconstrained and unstructured high-level preference concepts (e.g. interviews, design logbook, text description)  and the specific constrained and structured low-level features (e.g geometric shapes, CAD drawing, final reports) associated with those concepts. The main objective of the paper is focused on retrieving and generating preference concepts. The authors define preference as a subjective property that emerges from customer behavior, representing their desired choices. 
The model begins by semantically extracting customer preferences from their reviews. These extracted preferences are then subjected to part-of-speech (POS) tagging using semantic rules based on concepts outlined in a customer preference ontology and a preference lexicon.

Another related approach, discussed in \cite{tapucu} , aims to connect a preference model and an ontology-based database. The authors introduce a preference resource concept “property\_or\_class” within their preference model, which can be linked to the classes or properties defined in the ontology model. Their proposed preference model differentiates between interpreted and non-interpreted preferences. Interpreted preferences can be assigned an interpretation by evaluating their interpretation function. On the other hand, non-interpreted preferences are represented through an enumeration of a set of properties and classes selected from the ontology without any constraints.


\section{Scope}
\label{sec:scope}

In this work, our objective is to develop a preference ontology that can effectively address situation-dependent questions concerning user preferences. It is important to note that our focus lies in establishing the ontology itself, rather than learning the preferences. Therefore, for the purpose of presenting the ontology and its reasoning capabilities, we consider the preferences a-priori knowledge. However, in real-world scenarios, the required preferences could also be derived through learning algorithms. To showcase the supported reasoning of our proposed ontology, we have selected four competency questions that highlight its capabilities.
The proposed competency questions are:
\begin{enumerate}[label=\textbf{CQ\arabic*}]
    \item What are a user's preferences related to a particular situation? (E.g., what information do we have about a user's ordering of breakfast dishes they might prefer.)
    \item Which option would the user prefer from those that are about a given situation? (E.g., what would the user prefer to eat for breakfast.)
    \item Which preferences can be fulfilled by the agent? (e.g. we have no sugar, but other sweeteners that we can offer)
    \item Can preferences be inferred from other preferences? (E.g., preferring an effect, therefore preferring its cause)
\end{enumerate}

\section{Preference Ontology}
\label{sec:ontology}

\subsection{Ontological foundations}
We decided to extend the formal OWL
ontology \emph{Socio-physical Model of Activities (SOMA)} \cite{Bessler2021}.
Our reasons are three-fold. First, SOMA builds upon the foundational ontology \emph{DOLCE+DNS Ultralite (DUL)} \cite{DUL40}, which contains patterns to represent \emph{hypothetical worlds} conceptualized as \texttt{Situation}s and to model \texttt{Description}s thereof \cite{pisanelli2003ontology}. As we will explain in more detail in \Cref{subsec:pref}, these will be of utter importance for our model of preferences. Second, SOMA was recently extended to not only model physical experiences, for which it was originally designed, but to also capture metacognitive experiences \cite{Nolte2023}. It thus already contains concepts such as \texttt{Decision-Making}, \texttt{Option} and \texttt{Choice}, on which preferences are inseparably dependent. Third, SOMA contains an elaborate model of \texttt{Disposition}s. As shown later, this will be useful for the definition of preferences.

DUL, adopting a descriptive world view, consistently distinguishes \texttt{PhysicalEntities} from
\texttt{SocialEntities} that exist ``for the sake of [\dots] contextualizing or interpreting existing entities'' \cite{DUL40} such as \texttt{Description} and \texttt{Concept}.
Accordingly, a physical \texttt{Event} (an objective occurrence) \texttt{executes} an \texttt{EventType} (a subclass of \texttt{Concept} and thereby a subjective interpretation of said event), that \texttt{isDefinedIn} some \texttt{Description} such as an observer's \texttt{Narrative} or an emerging \texttt{Affordance}.
More concretely, a movement may \texttt{execute} a cutting \texttt{Task}.

DUL applies the above principle to other common ontology patterns such as \texttt{Role}s, which it also views as a subclass of \texttt{Concept} and therefore as defined by some \texttt{Description}.
Any object $x$ that -- objectively -- \texttt{isParticipantIn} some \texttt{Event} that \texttt{executes} a \texttt{Task} $y$ may be associated with a subjective \texttt{Role} linking to $x$ via \texttt{hasRole} and to $y$ per \texttt{hasTask}.
This allows for contextualizing entities (see, e.g., \cite{Fan2001} for details), such as differentiating a cutting tool from the object being cut, or \texttt{Option}s from the \texttt{Performer} and the \texttt{Choice}s in \texttt{DecisionMaking}.

\subsection{Preferences}
\label{subsec:pref}

We first discuss the agentive behaviors in which preferences manifest, i.e., the decision-making processes they guide.
%
{
SOMA views \texttt{Decision-making} as a \texttt{MentalTask} in which only an agent's mental representations (\texttt{InformationObject}s that play the roles of, e.g., \texttt{Premises} and \texttt{Conclusions}) are manipulated  (c.f.\ \cite{Nolte2023}). We can concretize that the \texttt{InformationObjects} behind \texttt{Premises} (\texttt{Conclusions}) contain information about the available \texttt{Options} (selected \texttt{Choices}). For the remainder of this work, however, we will abstract away from this pattern and only consider the \texttt{Options} and \texttt{Choices} as roles played by the considered entities during \texttt{Decision-Making}.}

Although some work exists on representing the comparative relationships between different options (c.f. \cite{sep-preferences}) we are unaware of any models of the options themselves that can answer questions such as, what is the ontological type of options, and how do they relate to other entities.
Such a model, if to be applied in flexible scenarios, would have to be of considerable complexity.
For illustration purposes, consider the following question:

\begin{quotation}
    ``Would you like coffee or tea?''
\end{quotation}

Assuming that neither coffee nor tea is prepared implies that the options are not concretely existing, but hypothetical by nature.
Notice also that, as per social norms, the question marks the beginning of a sequence of events in which the questioner prepares the chosen beverage and transfers it to the addressee (along with an exclusive claim of consumption), who eventually drinks it.
By these pragmatics, the options are not the hypothetical beverages or even the hypothetical possession thereof, but the hypothetical consumption.
This view is consistent with typical characterizations of preferences (c.f. \cite{sep-preferences}).
If this is to be captured by our representation of options, we have to view them as the roles played by hypothetical worlds -- in DUL terms, as \texttt{Description}s, \texttt{Situation}s may satisfy and which play the roles during decision-making.
These observations lead to the following definition:

\begin{definition}
    {\emph{Decision-making} is a mental task in which the performers select from a set of \emph{options}, which describe hypothetical worlds, a subset called \emph{choices}.}
\end{definition}

\begin{figure}[htbp!]
  \centerline{\includegraphics{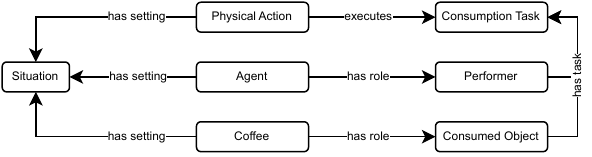}}
  \caption{ABox of a \texttt{Situation} in which an \texttt{Agent} consumes \texttt{Coffee}. Individuals are labeled with extending concepts.}
  \label{fig:coffee}
\end{figure}

DUL \texttt{Situation}s can be of arbitrary complexity: Encoded via a relation \texttt{hasSetting}, they can include, e.g., objects, actions, times and even other situations, which themselves can be linked to each other.
For example, the situation in which an agent consumes coffee may be modeled as seen in \cref{fig:coffee}.

We build upon this definition of decision-making to define preferences located as a quality of a bearer entity -- there is no preference that is not the subjective preference of somebody -- and emphasizes its role in decision making tasks performed by that bearer, in a manner similar to functionalist accounts of preference.

\begin{definition}
    A \emph{preference} is the disposition of its bearer to tend towards a particular subset of choices from a given set of options while performing decision-making tasks.
\end{definition}

Notice that the above definition does not make any assumptions on the method by which the options are evaluated and the final choices are selected.
Research has shown that human preferences are not necessarily rational \cite{Inza2008}, i.e, that they can contain cycles, and the cognitive mechanisms of decision-making are not completely understood yet.

To still capture the agents methodology underlying choice selection at least partially, we introduce a concept called \texttt{PreferenceDescription} that is a \texttt{Description} of a \texttt{Preference}.
In this exploratory paper, we focus on a simple, pragmatic representation of preferences via \texttt{PreferenceDescription}s as a partial order, similar to what has been proposed in preference logic (c.f.\ \cite{hansson2001preference}), although missing the associated modal semantics (notice that our approach's rich conceptualization of options as hypothetical worlds that can be described via \texttt{Situation}s instead far exceeds what is possible in preference logics that rely on propositional variables).

We argue that any such order underlying preferences is not between situations, but between descriptions thereof:
While the input of the above coffee-tea-scenario have been argued to be concrete situations of the decider consuming either, preferences are more general.
In our example, preferring tea over coffee is general in that sense that the agent values all situation in which he consumes the one drink over all similar\footnote{The compared situations must be similar except for the differences defining the choices: Otherwise, one might infer from an agents preference of coffee over tea that they would also prefer drinking coffee in agony over drinking tea while feeling splendid. Different solutions to this well-known problem exist \cite{liu2011reasoning}.}.
Ergo, such an order must be between descriptions (of situations):

\begin{definition}
    A \emph{preference order} is a partial order of descriptions (of situations).
\end{definition}

To represent such an order via OWL, we introduce not only the concept \texttt{PreferenceOrder}, but also \texttt{OrderedElement} and the relations \texttt{orderedBy} and $\leq$ (as well as the latter's inverse $\geq$).
The concept \texttt{OrderedElement} encapsulates \texttt{Description}s (of \texttt{Situation}s) and \texttt{orderedBy} exactly one \texttt{PreferenceOrder}.
The relations $\leq$ and $\geq$ have the obvious meaning of defining the order between the \texttt{OrderedElement}s.
This pattern allows to order the same \texttt{Description} (of \texttt{Situation}s) in multiple \texttt{PreferenceOrder}s as the ordering information is represented locally between \texttt{OrderedElement}s.

\begin{figure}[htbp!]
  \centerline{\includegraphics{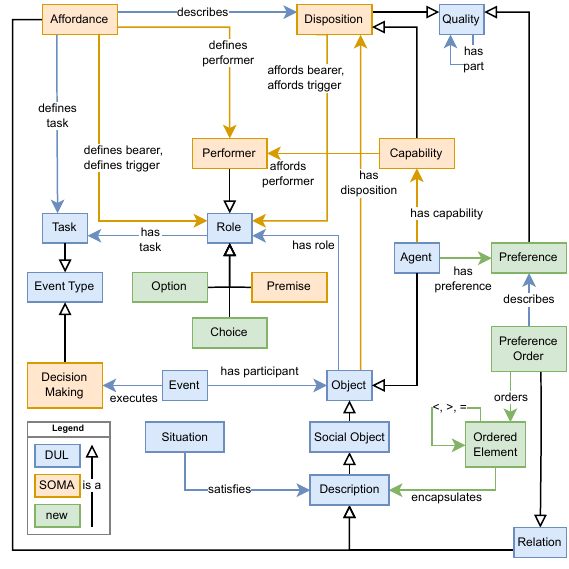}}
  \caption{Basic concepts and their relations of the proposed preference model (TBox); color-coded by defining ontology.}
  \label{fig:tbox}
\end{figure}

Together, the \texttt{PreferenceOrder} and the contextual nature of \texttt{Situation} allows to model very complex, conditional preferences; e.g., that someone prefers their spouse to prepare a meal on days where they work, but otherwise likes to cook themselves.
\Cref{fig:tbox} depicts the model as laid out in this section.

\section{Validation}

Using Konclude \cite{2014konclude} on a Ryzen 7 5800X CPU, 32GB RAM, and a GTX 3070 ti graphics card, initial classification of the current SOMA version, including the prefences ontology from section \ref{sec:ontology}, takes about 66ms, and subsequent DL queries take about 1ms each.

We hybridly evaluate the ontology by using the CQs from section \ref{sec:scope} and by giving reasoning and querying examples. Following the SABiO guidelines \cite{de2014sabio}, we use the listed formal queries as test-cases for the CQs. The queries are executed over example ABox-data with the preference ontology as the reasoning schema.

Figure \ref{fig:query_a} shows query A which is associated with CQ 1, asking for all preferences which a user has in a specific situation. Similarly, figure \ref{fig:query_b} shows query B which is associated with CQ 2, also the user and situation, but only the preference with the highest order, which reflects the user's preferred option for the situation.

\begin{figure}[htbp!]
\centering
\begin{minipage}{.48\textwidth}
  \centering
  \begin{lstlisting}
SELECT ?user ?sit ?pref WHERE {
    ?user a Agent;
          hasPreference ?pref.
    ?sit a Situation;
         satisfies ?desc.
    ?pref a Preference.
    ?desc a Description.
    ?elem a OrderedElement;
          encapsulates ?desc.
    ?ord a PreferenceOrder;
         orders ?elem.
         describes ?pref. }
    \end{lstlisting}
    \caption{Query A, associated with CQ 1}
    \label{fig:query_a}
\end{minipage}%
\begin{minipage}{.58\textwidth}
  \centering
  \begin{lstlisting}
SELECT ?user ?sit ?pref WHERE {
    ?user a Agent;
          hasPreference ?pref.
    ?sit a Situation;
         satisfies ?desc.
    ?pref a Preference.
    ?desc a Description.
    ?el a OrderedElement;
        encapsulates ?desc.
        NOT IN (SELECT ?e WHERE
                ?e a OrderedElement.
                ?f a OrderedElement;
                   greater ?e.) 
    ?ord a PreferenceOrder;
         orders ?elem.
         describes ?pref. }
    \end{lstlisting}
    \caption{Query B, associated with CQ 2}
    \label{fig:query_b}
\end{minipage}
\end{figure}

\noindent For space reasons, other queries and related machinery such as SWRL rules that are necessary for CQ 3 and 4 are not given in this paper but can be made available on request. 


\section{Conclusion}

We have introduced a formal ontology-based model for representing preferences. This model serves as a structured framework for capturing and organizing user preferences in the context of robot-human interactions. To evaluate our model, we formulated and examined relevant competency questions for the robot to understand and address user preferences in these interactions.

In future work, we will investigate the usage of learning techniques to incorporate learned user preferences alongside a-priori knowledge of specific user preferences. In previous research, we already implemented a model for interactions in our Ontology with the MOI ontology\cite{Nolte2023}, which provides a framework for logging episodic memories of past interactions. By extracting information from these episodic memories, we can learn user preferences. 
{We will furthermore look at how to automatically match given options to existing preferences, which is non-trivial: E.g., the situation of the user drinking tea is, by common sense, different from the situation in which the user drinks coffee. However, in the light of the open world assumption, such behavior has to be explicitly stated -- otherwise, one might come to the conclusion that a user might prefer drinking both simultaneously. It might be worth to look at how non-monotonic reasoning techniques could support such scenarios.  }

\bibliography{main}
\end{document}